# CheMLFlow: An Open-Source Platform for Cheminformatics and Materials Informatics Applications

Brendan Smith[1,†] Susana López-Moreno[2,3,4,†] Eric Dolores-Cuenca[5], Sangil Kim[2,4], Jose L. Mendoza-Cortes[6,7,*] Nijamudheen Abdulrahiman[1*]

[1]Kernfield Labs, London, United Kingdom.
[2]Department of Mathematics, Pusan National University, Republic of Korea.
[3] Humanoid Olfactory Display Center, Pusan National University, Republic of Korea.
[4] Industrial Mathematics Center, Pusan National University, Republic of Korea.
[5] Yonsei University, Republic of Korea.
[6] Department of Chemical Engineering & Materials Science, Michigan State University, United States.
[7] Department of Physics & Astronomy, Michigan State University, East Lansing, Michigan 48824, United States.

[*]Corresponding author.

[†]Contributed equally.

# ABSTRACT

CheMLFlow is an open-source platform for building and executing end-to-end, high-throughput, and agentic workflows for scientific and technological applications. CheMLFlow targets a common bottleneck in scientific machine learning development, where researchers often need to assemble data acquisition, curation, representation, model training, validation, screening, interpretation, and reporting into a reproducible pipeline, even when their primary research contribution concerns only one stage. CheMLFlow provides modular workflow components, ready-to-run reference pipelines, standardized artifacts, and evaluation outputs that reduce orchestration overhead and support benchmarking across methods and datasets. The platform is designed to be extensible, reproducible, and automation friendly, with pluggable representations and models, deterministic splits, explicit run artifacts, batch execution, and report generation. As scientific software increasingly moves toward agent assisted experimentation, CheMLFlow's configuration driven workflows and structured outputs also provide a practical interface for coding agents to help users construct experiments, inspect results, and summarize findings under human supervision. This article describes the system architecture, core workflows, and benchmarks that reach literature performance for quantum mechanical, physicochemical and bioactivity property prediction, and use cases involving time series datasets demonstrating applications beyond molecular chemistry datasets.

## Keywords

agent-assisted science, reproducibility, drug discovery, QSAR, explainable AI

# 1. INTRODUCTION

Machine learning (ML) and artificial intelligence (AI) are reshaping scientific and technological research by providing practical tools to understand, predict, and design complex molecular and material systems. Cheminformatics and materials informatics combine ML/AI with traditional computational chemistry, statistical modeling, and domain specific representations for data driven prediction and design of molecules and materials. Cheminformatics is widely used in computer aided drug discovery and AI aided drug discovery, including virtual screening and synthesis route/process optimization of active pharmaceutical ingredients, while materials informatics is actively pursued for next generation energy materials, catalysts, semiconductors, superconductors, magnets, and functional materials.[1–10] Quantitative structure activity/property relationship (QSAR/QSPR) modeling forms an essential part of both fields and is used for activity/property prediction as well as inverse design. Recently, deep QSA(P)R methods have advanced rapidly through graph neural networks, molecular foundation models, and generative AI.[11–16] At the same time, agentic methods are increasingly being incorporated into scientific workflows, as demonstrated by tools such as ChemCrow,[17] Co-Scientist,[18,19] and El Agente,[20,21] among others.[22] Despite this progress, many available tools remain focused on individual modeling or design tasks, and improvements are still needed in end-to-end workflow support, reproducibility, and systematic benchmarking across methods and datasets.[23–26]

For deep QSA(P)R modeling and beyond, scientific ML/AI workflows usually require several connected stages as demonstrated in **Figure 1**: (1) data generation or data acquisition, (2) data curation, (3) exploratory data analysis (EDA), (4) data screening based on desired or undesired properties, (5) data representation, (6) data engineering, (7) model training, (8) validation, (9) domain applications, and (10) the generation of reports. Most research and technical programs focus on one or a few of these stages rather than optimizing the entire workflow. Nevertheless, complete studies often require researchers to implement nearly all of them, even when their primary contribution concerns only a new representation, model, dataset, or application. This repeated orchestration effort slows iteration, complicates fair comparisons, and makes reproducibility difficult.

Here, we introduce CheMLFlow, an open-source platform for developing and applying cheminformatics and materials informatics workflows. CheMLFlow is designed as a modular and scalable system with ready-to-use workflows for molecular property prediction tasks relevant to cheminformatics and materials informatics, including quantum mechanical, physicochemical, fuel relevant, and ADMET property prediction. The platform is intended to help non-expert users implement complete workflows to domain problems without managing every technical detail, while also allowing expert researchers to focus on individual stages such as data generation, representation, or ML/AI model development. CheMLFlow enables leaderboard style benchmark summaries through high throughput testing of complete workflow configurations, in which the dataset, curation, representation, split strategy, preprocessing choices, model, and analysis outputs are treated as a single reproducible benchmark unit within a common workflow structure. This is important because model performance in practice is often problem dependent. Therefore, a traditional method such as random forest or XGBoost can outperform a more complex deep learning model for some datasets, while graph-based or SMILES-native representations may be advantageous in other settings. CheMLFlow automatically creates artifacts with the information required to reproduce the experiments. Although the primary demonstrations focus on molecular property prediction, the same node-based workflow abstraction can support other scientific data modalities. This extensibility is illustrated below through a time series forecasting module for chaotic dynamical systems. In addition, CheMLFlow's configuration driven execution, structured run artifacts, and agent facing skills make it well suited for agent assisted scientific computing where coding agents can help users install the software, construct valid experiments, inspect outputs, run analyses, and summarize results under human supervision, while scientific interpretation and final decisions remain with the researcher.

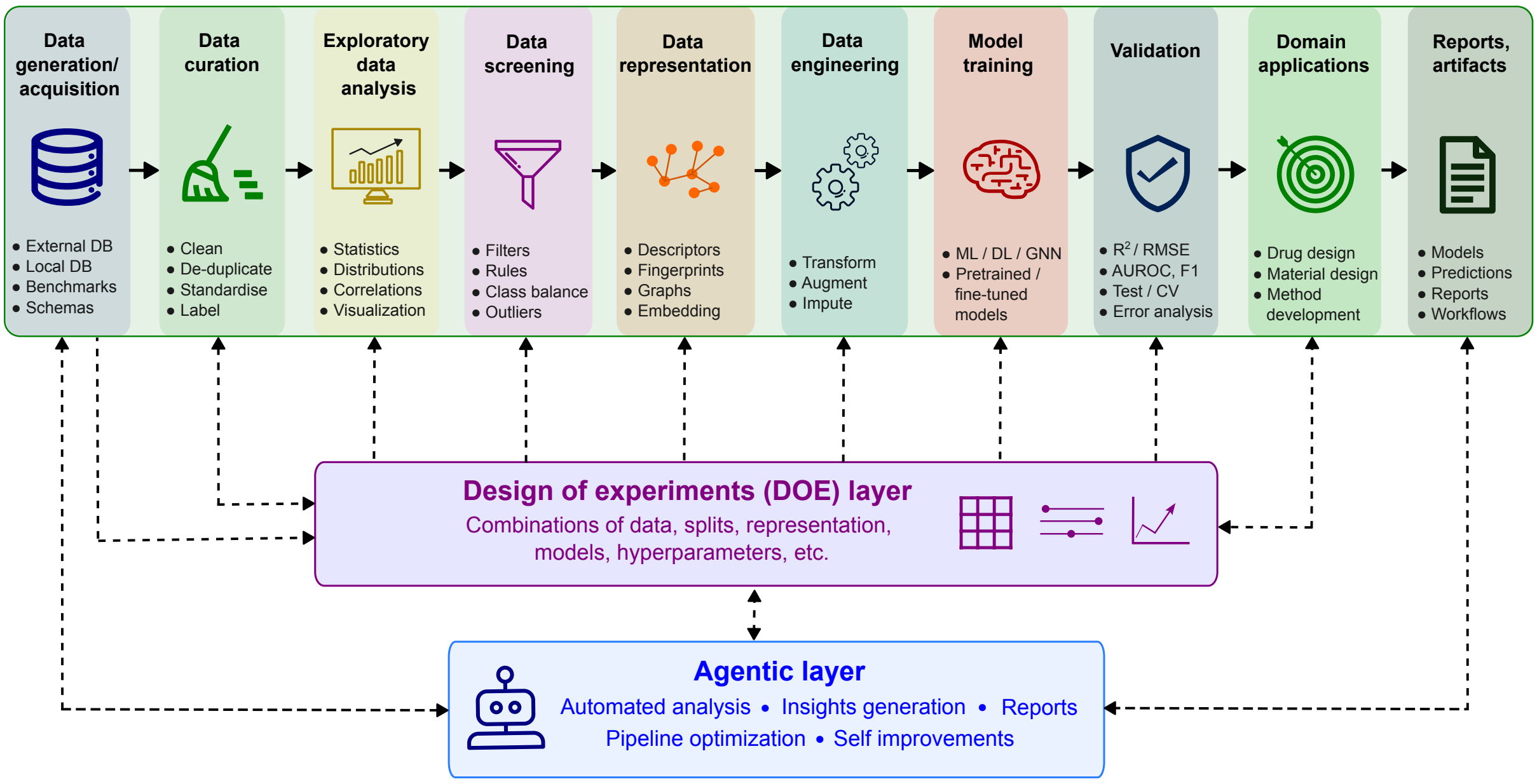

**Figure 1**. A schematic demonstrating three layers of CheMLFlow to construct workflows. (i) A core workflow builder layer (on the top) for constructing an end-to-end (or node-to-node) pipeline of training and prediction starting from data generation/acquisition to reports with specific options chosen at each node. (ii) An optional design of experiments (DOE) layer for high throughput experimentation of model combinations, where multiple options are chosen at each node of interest to produce a leaderboard of performance metrics. (iii) An optional agentic layer, that can be used to set up workflows and DOEs, generating reports, and for optimizing workflows through prompts either in plain non-technical conversation language or specific technical comments to guide and secure task executions.

# 2. SOFTWARE AND METHODS

## 2.1. System Design and Workflow Specification

CheMLFlow is designed as an open-source platform to implement ML/AI workflows (**Figure 1**) using a command line interface (CLI) and/or an agent. A workflow is designed as composed of individual nodes or stages, where each node based run writes the resolved configuration, status, executed pipeline nodes, split definitions where applicable, and produced artifacts such as metrics, reports, plots, and model files where supported. CheMLFlow provides multiple options for each node starting from data acquisition to validated models and reports. CheMLFlow can generate parity and split metric plots for regression, ROC/PR/confusion plots for classification, permutation importance and SHAP explanations where supported, EDA outputs including an interactive web interface for specific applications, and DOE aggregate analysis tables from recorded run artifacts. CLI format can be used for local execution, batch scripts, HPCC schedulers, and external orchestration tools. Scientific comparisons often depend on more than the model alone. Dataset curation, molecular representation, split strategy, preprocessing, random seed, and failure handling can all impact the performance. CheMLFlow therefore includes a design-of-experiments (DOE) layer to compare workflow configurations across folds, seeds, and parent configurations with less manual bookkeeping. Further, the structured artifacts produced by CheMLFlow makes the platform well suited for human-on-the-loop agentic workflows. In addition to user facing documentation, CheMLFlow provides agent facing skills that help coding agents install the software, construct valid configurations, launch runs, locate outputs, run analysis, and summarize results. Overall, the emphasis is on making scientific ML runs explicit, repeatable, comparable, and inspectable, while leaving enough flexibility for researchers to adapt individual components to new datasets, representations, and modeling questions.

## 2.2. Workflow Model

Workflows are configured through a single YAML or JSON run configuration and executed from the command line. The configuration defines the dataset source, task type, target column, pipeline nodes, curation settings, split strategy, molecular representation, preprocessing options, model type, training settings, and output locations (**Figure 2**). This makes the workflow explicit and version controllable. Rerunning the same resolved configuration with the same random seed and compatible software environment should reproduce the same pipeline structure and split definitions. A single workflow run is executed by pointing CheMLFlow to one configuration file. During execution, each configured node reads the artifacts produced by earlier nodes and writes its own outputs to the run directory. For example, a supervised property prediction workflow may ingest a local CSV file, curate valid SMILES records, generate descriptors or fingerprints, create train/validation/test split indices, preprocess feature matrices, train a model, and write performance metrics and predictions. A completed run produces a directory containing the resolved run configuration, run status, split metadata where applicable, model metrics, predictions, plots, and model artifacts where supported.

A run is expected to produce a minimal, consistent set of artifacts such as:

1. DOE specifications and/or generated individual run configuration files.
2. Files capturing workflow state, including run status, timestamps, config path, run directories, nodes, configuration hashes, failure metadata if applicable, and git hash.
3. Saved train/validation/test row identifiers and split metadata generated.
4. Model specific metrics files capturing task appropriate metrics for each completed run. DOE analysis aggregates successful child runs into parent level summary tables.
5. Model specific prediction files capturing true and predicted values, and probabilities or scores for classification produced by the selected model workflow.
6. Serialized model files and training parameters, if supported by the selected model.
7. Plots such as parity and split metric plots for regression, ROC/PR/confusion plots for classification, and permutation importance or SHAP summaries where supported.

8. Analysis outputs such as aggregate CSV/JSON summaries and figures generated from DOE or run artifacts where the analysis workflow is executed.

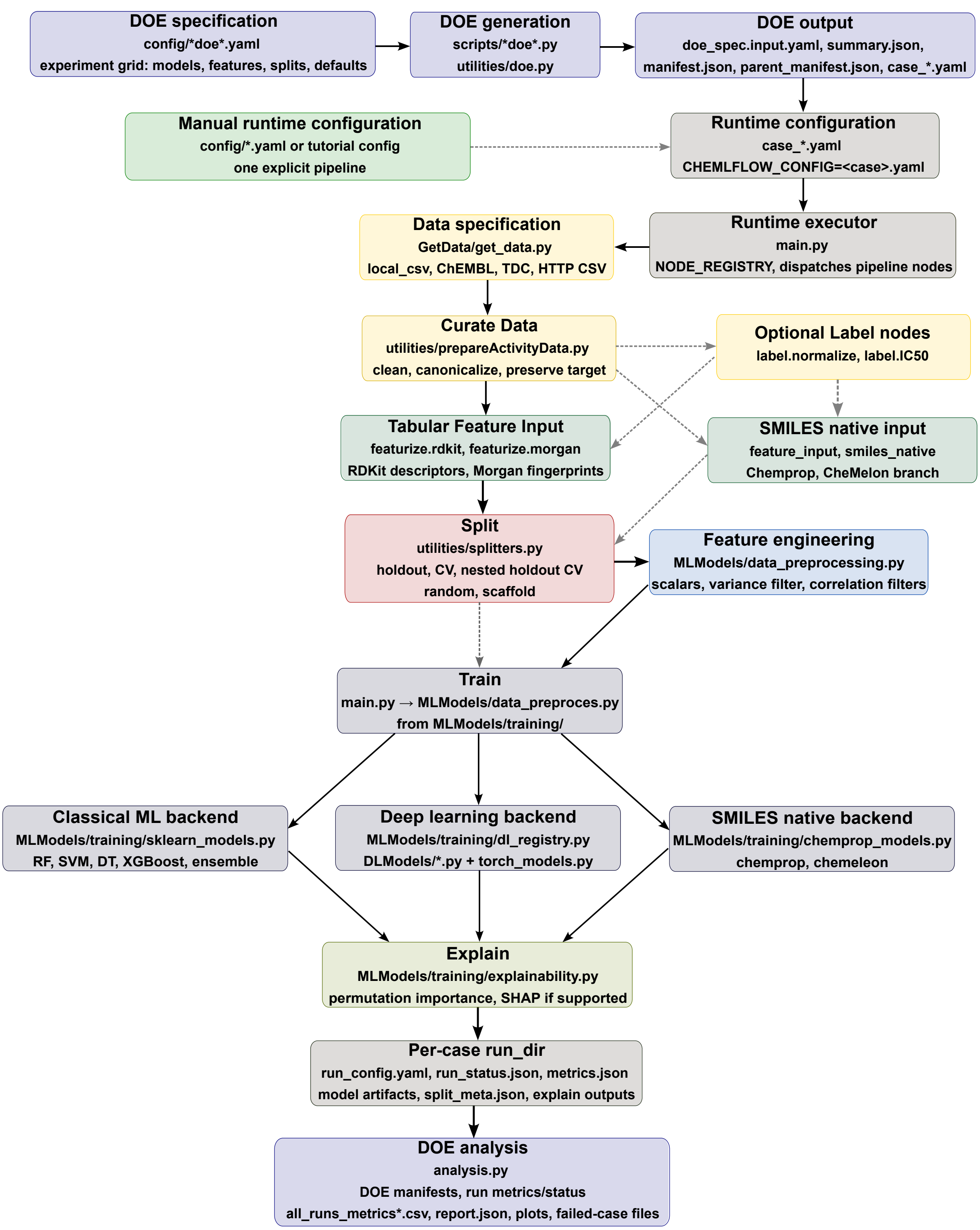

**Figure 2.** CheMLFlow tree structure presenting definitions of DOE guided high throughput runs of model combinations and individual config driven workflow runs along with options available at individual nodes. The scripts and files defining options at node level are also shown.

This artifact contract makes benchmark results auditable and allows tables and figures to be regenerated from recorded outputs rather than manually assembled.

## 2.3. Benchmark DOE Protocol

For a systematic comparison of workflow choices, CheMLFlow supports DOE runs in which multiple representations, split strategies, preprocessing settings, and models are evaluated under a shared protocol, defined through explicit configuration files. Specifically, CheMLFlow accepts a higher level DOE specification, which expands a grid or model search definition into many concrete single run configuration files. DOE cases may be executed manually, by local scripts, or by a compute cluster scheduler. The execution backend is not part of the scientific definition of the experiment. The scientific unit is the resolved case configuration and its recorded artifacts. After execution, the analysis step aggregates successful child runs into parent level summaries and records failed child runs separately so that invalid model outputs are not converted into misleading metrics.

Each DOE case is executed as an independent run and writes its own status, configuration, split metadata, metrics, predictions, and model artifacts where supported. Failed cases are retained as failed runs rather than converted into artificial metric values. During analysis, successful runs are aggregated into parent level summaries, while failed or incomplete runs are counted separately. This prevents invalid outputs, such as nonfinite predictions, from biasing model comparison tables. This DOE protocol makes the benchmark unit a complete workflow configuration rather than only a model name. As a result, performance tables can be interpreted as comparisons over dataset, representation, split strategy, preprocessing, and model choices under a reproducible execution contract.

## 2.4. Documentation and Agent Facing Skills

CheMLFlow includes agent facing skills in the form of short, versioned operating guides that help coding agents generate valid configurations, launch runs, locate outputs, run analysis, and explain results back to a human user. These skills make routine orchestration less fragile by giving agents clear instructions for working with configurations, manifests, metrics tables, plots, and run status artifacts while leaving interpretation and final decisions with the human researcher.

# 3. APPLICATIONS AND RESULTS

## 3.1. Application of CheMLFlow to Bioactivity Prediction

We demonstrate the use of CheMLFlow for DOE-driven, end-to-end ML/AI training for bioactivity (IC50) prediction tasks using datasets retrieved from ChEMBL database in the initial step of the workflow run.[27] In each study, DOE specification was defined by a ChEMBL target protein/assay identifier together with node choices including data retrieval, data cleaning and curation, molecular featurization, data splitting strategy, scaling, ML/AI or pre-trained model family, optional SHAP explainability where supported, and the performance metrics to be reported. Specifically, literature comparable baseline models were developed for four ChEMBL target datasets namely PDE4B/CHEMBL275,[28] EGFR/CHEMBL203,[29] hERG/CHEMBL240,[30] and ALK-5/CHEMBL4439.[31] The DOE configuration files and execution/analysis scripts were prepared and executed through an author reviewed, agent-assisted scripting workflow using CheMLFlow with Codex using GPT-5.5 as the agent. This approach was used to train models and representation combinations that were comparable to those reported in the corresponding literature studies, while recognizing that the current CheMLFlow implementation does not include every model architecture, descriptor set, or optimization protocol used in those reports. Despite this intentionally baseline oriented setup, CheMLFlow produced strong performance across the selected ChEMBL targets, including competitive results for PDE4B and EGFR, a practically useful hERG baseline, and an ALK-5 model that exceeded the reported literature DNN validation $R^2$ under random 5-fold cross-validation.

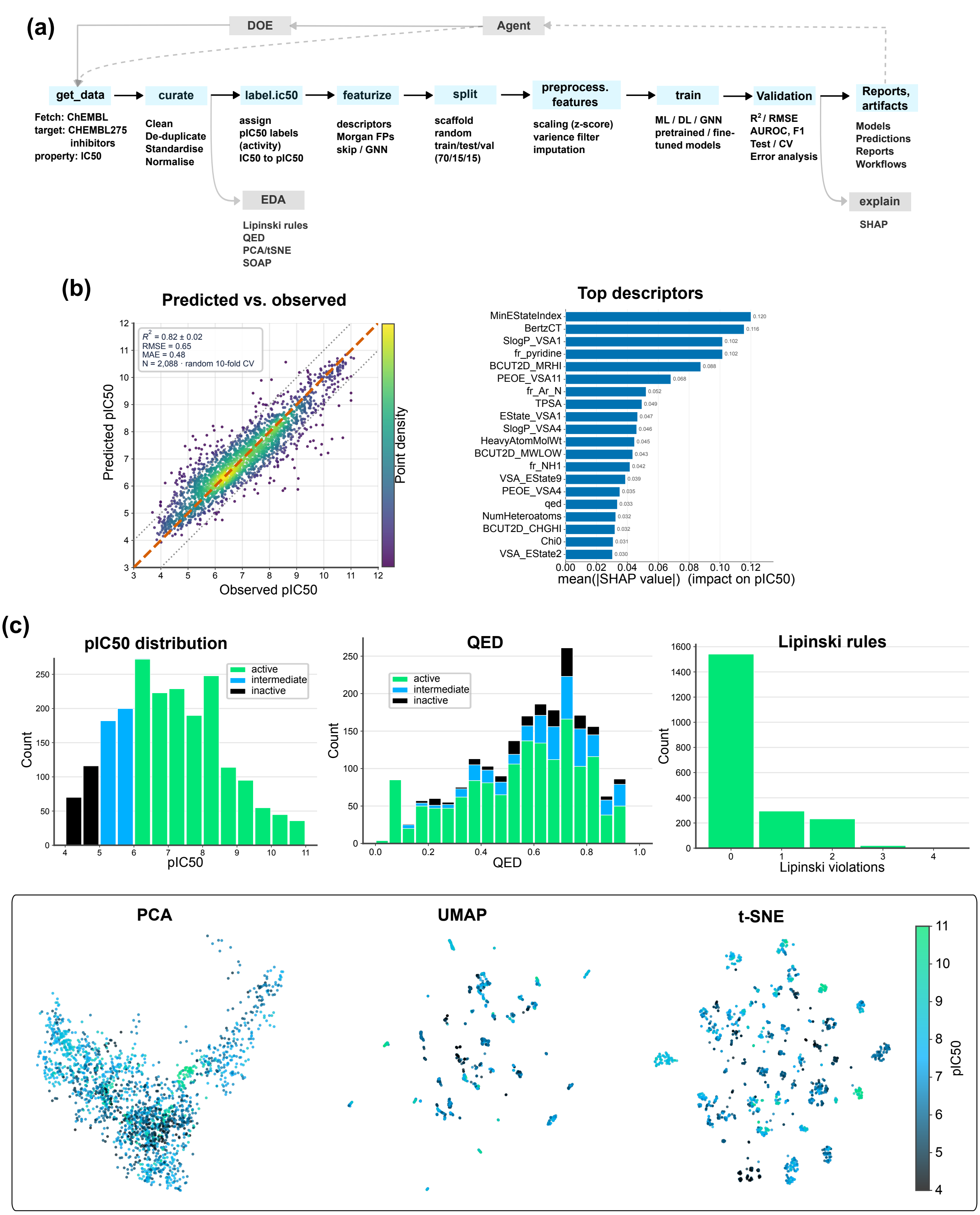

**Figure 3**. (a) Flowchart of a typical end-to-end workflow to train and predict bioactivity values using IC50 data retrieved from ChEMBL dataset, (b) model performance plot for an ensemble model trained on pIC50 data (for target ChEMBL275) and a SHAP analysis plot generated, (c) optional EDA nodes plots of pIC50 distributions, QED, Lipinski rule violations, PCA, UMAP, and t-SNE.

As summarized in **Figure 3**, the ChEMBL bioactivity workflow begins with the retrieval of IC50 records, followed by molecular curation, endpoint standardization, and conversion of IC50 values to pIC50. The curated molecules are then passed through optional exploratory analysis nodes, including pIC50 distribution analysis, molecular property visualization, dimensionality reduction plots such as PCA or UMAP, and drug likeness analyses based on Lipinski rules[32] and quantitative estimate of drug likeness (QED).[33] When enabled, CheMLFlow supported molecular map visualizations can also be used to inspect how molecular structures, target activity values, and chemical property distributions are organized across the dataset. Optionally, further extensive details of datasets can be accessed through an interactive web interface. These analyses are useful before model training because they expose data imbalance, activity range limitations, structural clustering, and potential chemical series effects that may influence model performance.

After curation and exploratory analysis, CheMLFlow generated molecular representations and executed DOE defined model training workflows using Morgan fingerprints and RDKit descriptors as the main feature representations.[34] Model performance was evaluated using both random and scaffold 5-fold cross-validation. Random cross-validation estimates interpolation performance within the curated chemical space, whereas scaffold cross-validation provides a more conservative estimate of generalization to new chemical series. Reporting both split strategies is important for these studies because structurally related analogs can otherwise appear in both training and test folds, leading to overly optimistic estimates of model performance.

Across the four ChEMBL target studies, CheMLFlow produced literature adjacent baseline models, with the strongest results observed for PDE4B, EGFR, and ALK-5. For PDE4B/CHEMBL275, an ensemble model with Morgan fingerprints achieved a random-CV $R^2$ of 0.81 and MAE of 0.51, approaching the reported AutoML ensemble $R^2$ of 0.85. Further, an ensemble model with a 10-fold CV led to $R^2$ of 0.82. For EGFR/CHEMBL203, a random forest model with Morgan fingerprints achieved $R^2 = 0.70$ and MAE = 0.5, close to the reported random-forest benchmark of approximately $R^2 = 0.72$. For ALK-5/CHEMBL4439, random forest with Morgan fingerprints achieved $R^2 = 0.69$ and MAE = 0.47, exceeding the reported DNN validation $R^2 = 0.66$. The hERG/CHEMBL240 study gave a more conservative but still useful baseline, with SVM/Morgan achieving $R^2 = 0.52$ and MAE = 0.45, below the optimized Pred-hERG 5.0 regression R2 = 0.61 and RMSE = 0.48 but in a similar absolute-error range. Scaffold-CV performance was lower, as expected, but remained informative. PDE4B retained $R^2 = 0.71$, while EGFR, ALK-5, and hERG gave scaffold-CV $R^2$ values of 0.58, 0.56, and 0.36, respectively.

These comparisons were not intended as exact reproductions of the corresponding literature models, since ChEMBL version, assay filtering, descriptor choices, hyperparameter optimization, and split definitions can all affect final performance. Instead, they demonstrate that CheMLFlow can rapidly generate transparent, auditable, and literature comparable baseline studies from public bioactivity data. Each study was defined through explicit workflow factors, including target identifier, endpoint transformation, representation, preprocessing, split strategy, model family, cross-validation protocol, and output metrics, while CheMLFlow handled curation, featurization, training, metric aggregation, and artifact organization. Having established this workflow for ChEMBL-derived pIC50 prediction, we next evaluated whether the same DOE-guided strategy generalizes beyond bioactivity to electronic, physicochemical, ADME, receptor activity, and fuel relevant molecular property benchmarks.

## 3.2. Benchmarking Molecular Property Prediction workflows with CheMLFlow

**Table 1**. DOE-guided benchmarking of models for six datasets in predicting ADME, quantum mechanical, and physicochemical properties. Task, size of the dataset, regression/classification type of task, best reported models, performance metrics, literature value, and top performing model in a single shot DOE-guided experiment.

| Dataset | Task | Size of dataset | Type | Best Models | | Metrics | Performance | |
|---|---|---|---|---|---|---|---|---|
| | | | | Reported | CheMLFlow (single shot DOE run) | | This study | Literature |
| QM9[35] | Electronic: energy gap | 133885 | Reg. | Chemprop | Chemeleon | MAE | 0.0043 | 0.0047[36] |
| Pgp[37] | ADME: P-gp | 1275 | Cls. | MLP [38] | Random forest | AUC-ROC | 0.95 | 0.95[38] |
| ARA[39] | ADME: Androgen receptor | 842 | Cls. | DeepAR | Ensemble | AUC-ROC | 0.95 | 0.95[39] |
| Flash[40] | Physchem: Flash point | 632 | Reg. | Ensemble | DL-deep, Ensemble | $R^2$ | 0.92 | 0.94[40] |
| YSI[41] | Physchem: Sooting index | 442 | Reg. | Bayesian linear regression | Ensemble | MAE | 19.17 | 22.3[41,42] |
| PAH[43] | Physchem: log | 55 | Reg. | Linear regression | DL-GRU | $R^2$ | 0.96 | 0.96[43] |

CheMLFlow's implementation of DOE guided benchmarking of models was studied for property prediction tasks using 6 small molecule datasets that are publicly available and commonly used for validating ML/AI models. The datasets, properties of interest, regression or classification type, performance metrics, and other information are presented in Table 1. These datasets were chosen following a recent work by Burns and Green,[42] and they represent a range of molecular systems and some of the common tasks related to property predictions in chemistry and materials science such as ADME, quantum mechanical electronic structure, physicochemical and fuel properties. The size of datasets varies from 55 (PAH) to 133,885 (QM9) and contains simple aliphatic/aromatic hydrocarbons to advanced pharmaceutical ingredients and molecules of astrophysical importance. Previous ML/QSAR studies using these datasets allowed for a direct comparison of the performance of the methods here (**Table 1**).

A total of 136 aggregated parent jobs × 5 CV folds = 680 child jobs were run for a given dataset resulting in a total of 4080 child jobs run for all six datasets studied. For a given dataset, 11 models × 2 feature inputs (RDKit descriptors and Morgan fingerprints) × 3 scalers (minmax, standard, none) × 2 split strategies (random, scaffold) were performed on tabular descriptors or fingerprint data leading to 11 × 2 × 3 × 2 = 132 aggregated runs. Further, 2 SMILES-native Chemprop and CheMeleon models combined with random/scaffold splitting adds to 4 aggregated runs. ML models studied are decision trees, SVM, RF, XGboost, and ensembles; and DL models studied are DL-simple, DL-deep, DL-autoregression, DL-gated recurrent unit, DL-tab transformer, and DL-residual MLP. Chemprop[25] and CheMeleon[44] were evaluated as SMILES-native graph neural network baselines within the CheMLFlow workflow. Unlike the tabular models, these models did not use RDKit descriptors or Morgan fingerprints. Instead, curated canonical SMILES strings were passed directly to the Chemprop molecular graph pipeline, where molecules were converted into graph datapoints and featurized using Chemprop's molecular graph featurizer. For the Chemprop baseline, a message passing neural network was trained from scratch using a bond-message-passing encoder, mean aggregation, and a regression feed-forward prediction head. For CheMeleon, the same Chemprop training interface was used, but the message passing encoder was initialized from a pretrained CheMeleon checkpoint. In the runs reported here, the pretrained encoder was frozen and only the downstream regression head was trained, providing a foundation model baseline for molecular property prediction. Both Chemprop and CheMeleon used the same curated molecule sets and the same cross-validation split definitions as the other CheMLFlow models. Impacts due to random and scaffold splits and 5-fold cross-validation were evaluated separately. For each fold, a validation subset was drawn from the training data for neural network training and checkpoint selection. Performance was reported as mean and standard deviation of test-set metrics across the five folds.

Our DOE approach already trained workflows of model combinations that matched the performance metrics of literature SOTA methods. However, the goal here is not to test all possible models implemented in CheMLFlow to come up with best performing model, rather show the usefulness of the approach to quickly implement models, test workflows, and explore combinatorial spaces. Further, CheMLFlow implements optuna,[45] which can be used for optimizing model performance through hyper parameter optimization and regularization, which is not attempted here.

Selected plots of performance metrics corresponding to unique model combinations for a regression problem (QM9) and a classification problem (pgp) are presented below in Figure 4. The experiments revealed following observables. Overall, across the datasets and within DOE defined model space, the performance metrics are strongly influenced by the featurization, model class, and split strategy, as expected. RDKit based descriptors along with classical ML models such as random forest, xgboost, and ensemble methods yielded overall better performance than deep learning models for specific datasets. However, modern graph-based methods get better for larger datasets, and a further HPO is expected to improve the performance of all models. Random splitting led to better performing model combinations than those based on scaffold splitting strategy. However, the latter is recommended for transferable models for practical applications.

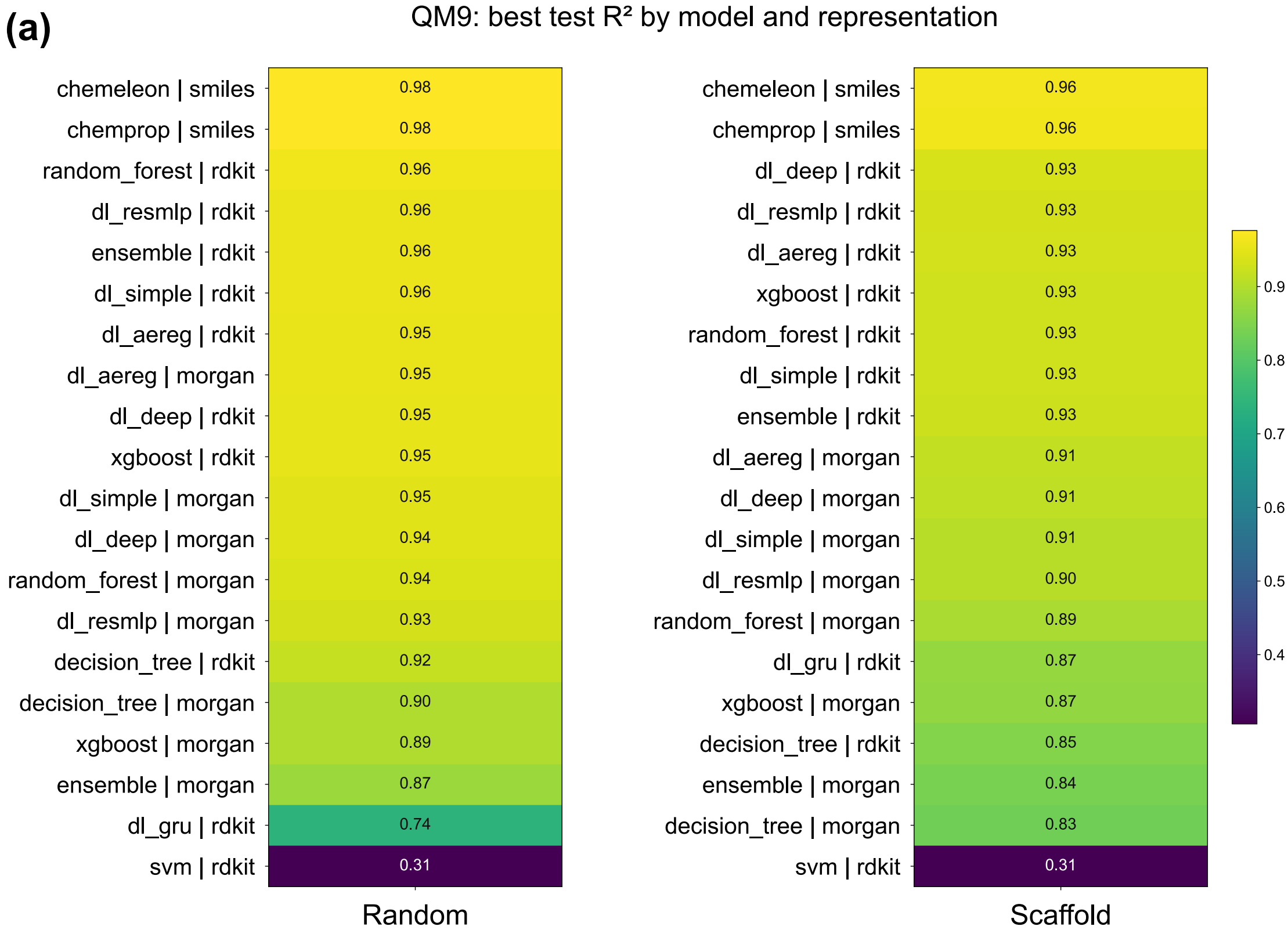

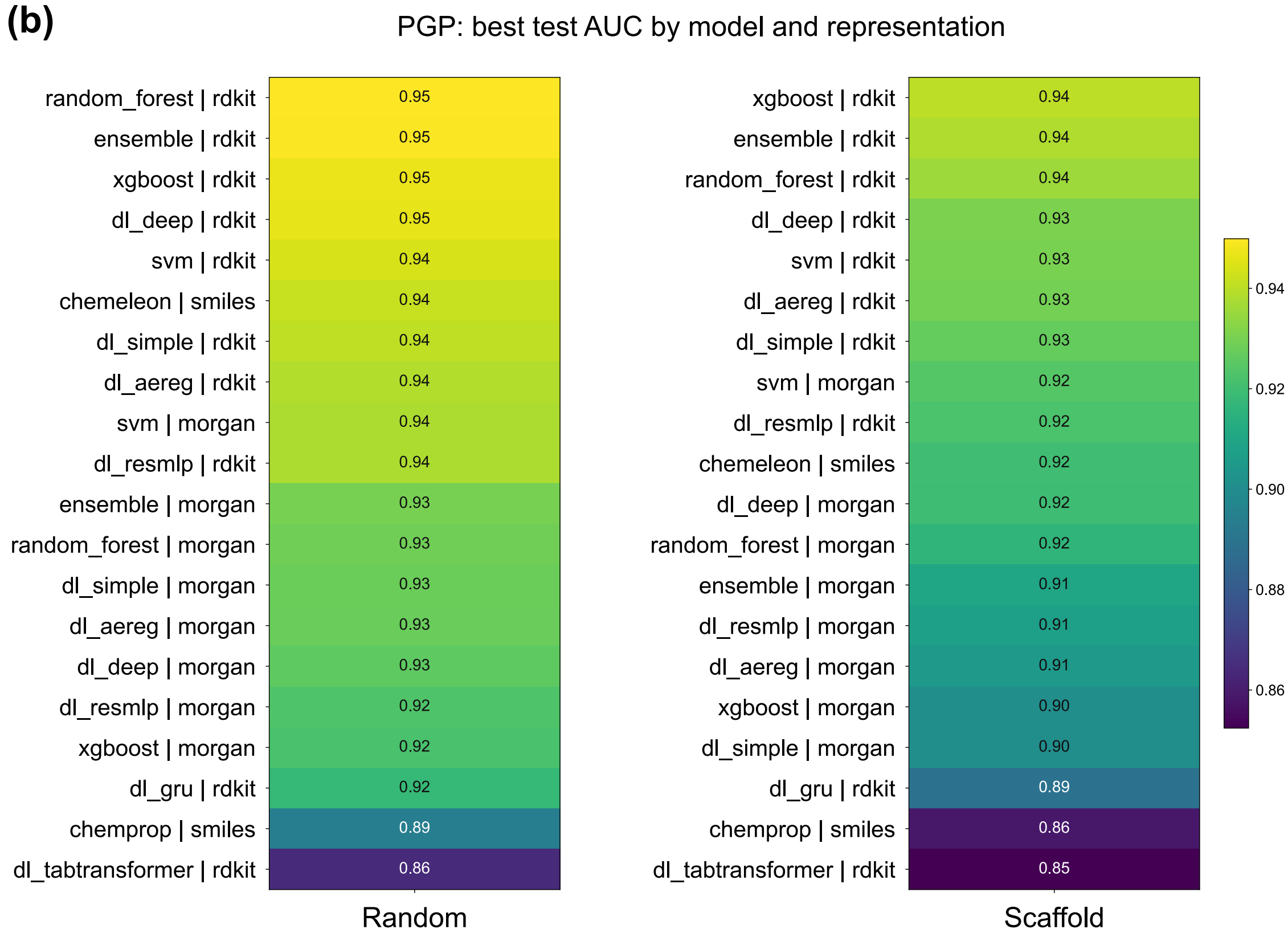

**Figure 4**. Performance of model combinations in predicting the (a) HOMO-LUMO gaps (regression task) and (b) Pgp (classification task) using QM9 and Pgp datasets, respectively. All values are means over the five folds.

Promisingly, scaffold splitting still retained good performance for most data sets except when the intrinsic heterogeneity of specific datasets, reduced the test $R^2$ values. An inspection of molecular structures of train, test, validation data support this observation. Further, the explainability based on SHAP analysis was enabled for model combination that used RDkit descriptors.

## 3.3. Agent Assisted Workflow Evaluation

CheMLFlow presents agent skills, as concise operating guidance for common points of failure, such as writing valid configuration files, choosing supported model and featurization combinations, preserving data splits, and auditing results. We evaluated agent skills on a binarized QuantumScents odor prediction task.[46] QuantumScents is a multilabel dataset in which molecules can be associated with multiple odor descriptors. We constructed a one-*vs*-rest task for the fruity odor label, assigning molecules annotated as fruity to label 1 and all other molecules to label 0. We used Codex powered by GPT-5.5 with extra high reasoning effort as the agent for two experiments; (1) the agent completes the task without CheMLFlow, (2) the agent uses CheMLFlow, including the repository documentation and CheMLFlow agent skills. Each experiment started with a clean working state so that the agent does not inherit cached outputs, previous configuration choices, or generated result files. CheMLFlow repository was installed from source and the python environment was activated before running the workflow. The CheMLFlow skills were installed before the agent begins the task, using Codex's local skill installer as, '*$skill-installer install CheMLFlow skills*'. We note, other coding agents might require a different mechanism for installing or exposing agent facing workflow instructions. After installation, the skills were made available in a fresh agent session before the task prompt is issued. The following prompt was used in both experiments.

*Prompt: Can you use this dataset to find the best model for predicting whether a molecule has a fruity odor from its structure?*

**Experiment 1** (Agent only baseline in absence of CheMLFlow): The agent completed the task as a standalone Python benchmark. It selected XGBoost with ECFP4 plus RDKit descriptors as the best model, with a 5-fold cross-validated ROC-AUC of 0.88 and a held-out ROC-AUC of 0.87.

**Experiment 2** (CheMLFlow with agent skills): The agent found the task as a scaffold 5-fold cross-validation study for the fruity odor prediction. CheMLFlow DOE compared Morgan fingerprints, RDKit descriptors, and SMILES-native Chemprop input across compatible model families. This comparison was scientifically useful because Chemprop is the message passing model family used in the original QuantumScents study, allowing the workflow to include a literature relevant neural baseline rather than limiting the comparison to descriptor-based models. DOE defined 9 distinct model/feature comparisons, and evaluated across 5 scaffold folds, producing 45 individual training runs. CheMLFlow also generated 30 additional cases that were skipped for expected model-feature incompatibilities. All 45 valid runs completed successfully, and the analysis audit confirmed 9 complete aggregate comparisons with no failed jobs, no missing metrics, and complete split diagnostics.

Following upon experiment 1 that identified ECFP4 plus RDKit descriptors as the strongest standalone representation under stratified random validation, we expanded the CheMLFlow DOE to make the comparison more direct. The expanded DOE added the combined ECFP4 plus RDKit descriptor feature branch and added stratified random, cross-validation alongside scaffold cross-validation. This expanded DOE defined 26 distinct model/feature/split comparisons. Because each comparison was evaluated with a 5-fold cross-validation, the expanded study produced 130 individual training runs. CheMLFlow also generated 70 additional cases that were skipped for expected model-feature incompatibilities. The final analysis audit confirmed all 130 valid runs completed successfully, producing 26 complete aggregate comparisons with no failed jobs or missing metrics. The expanded run reproduced the original scaffold only CheMLFlow results exactly for the 9 overlapping scaffold comparisons in the same local environment. The maximum absolute difference in ROC-AUC and AUPRC between the original scaffold only run and the expanded run was 0.0. Thus, the expanded DOE did not alter the earlier CheMLFlow findings; it extended them with an ECFP4 plus RDKit branch and a stratified random validation branch.

Under random, stratified validation in the expanded Experiment 2 run, the best CheMLFlow model was CatBoost with RDKit descriptors, with ROC-AUC 0.88. The best CheMLFlow model using the Experiment 1 - style ECFP4 plus RDKit representation was the RF/XGBoost ensemble, with ROC-AUC 0.88. These values closely matched the standalone Experiment 1 result of 0.88. Under scaffold validation, the best expanded CheMLFlow model was the RF/XGBoost ensemble with ECFP4 plus RDKit descriptors, with ROC-AUC 0.83. Within the original scaffold only search space, the best model remained CatBoost with RDKit descriptors, with ROC-AUC 0.83. Chemprop achieved ROC-AUC 0.81 under scaffold validation. The results therefore support two complementary conclusions. First, under stratified random validation, CheMLFlow found models that matched the standalone agent. Second, CheMLFlow improved the scientific quality of the experiment by making the validation strategy explicit and auditable. The best ECFP4 plus RDKit model dropped from approximately 0.88 under stratified random validation to 0.83 under scaffold validation. This approximately 0.05 ROC-AUC reduction is not a failure of CheMLFlow. Rather, it is a useful finding that the standalone benchmark was not designed to reveal.

## 3.4. Application to Time Series Forecasting

Time series are important in chemistry because many chemical and biochemical processes, such as reaction kinetics or molecular dynamics trajectories, evolve over time. Forecasting these data extends CheMLFlow beyond static regression and classification tasks, enabling the prediction of future system behavior, including in nonlinear or chaotic dynamical settings.

In addition to the ML and DL models discussed above, CheMLFlow also supports time series forecasting. It includes forecasting of chaotic times series through the Adaptive NVAR model.[47] CheMLFlow's implementation of a reported benchmark gives the results in **Table S1**. We start with the Mackey-Glass dataset, a synthetic time-series dataset generated by the following Mackey-Glass delay differential equation:

$$\frac{dx(t)}{dt} = -bx(t) + \frac{ax(t-\tau)}{1 + x(t-\tau)^n},$$

where the parameters $a = 0.2,\ b = 0.1,\ n = 10$ come from the classical Mackey-Glass model, while $\tau = 17$ is a widely used benchmark for generating chaotic Mackey-Glass time series. The resulting dynamics are shown in **Figure 5a**, including the dynamics with additional synthetic Gaussian noise. **Figure 5b** illustrates the non-overlapping 5 window forecast by Adaptive NVAR for the different levels of Gaussian noise.

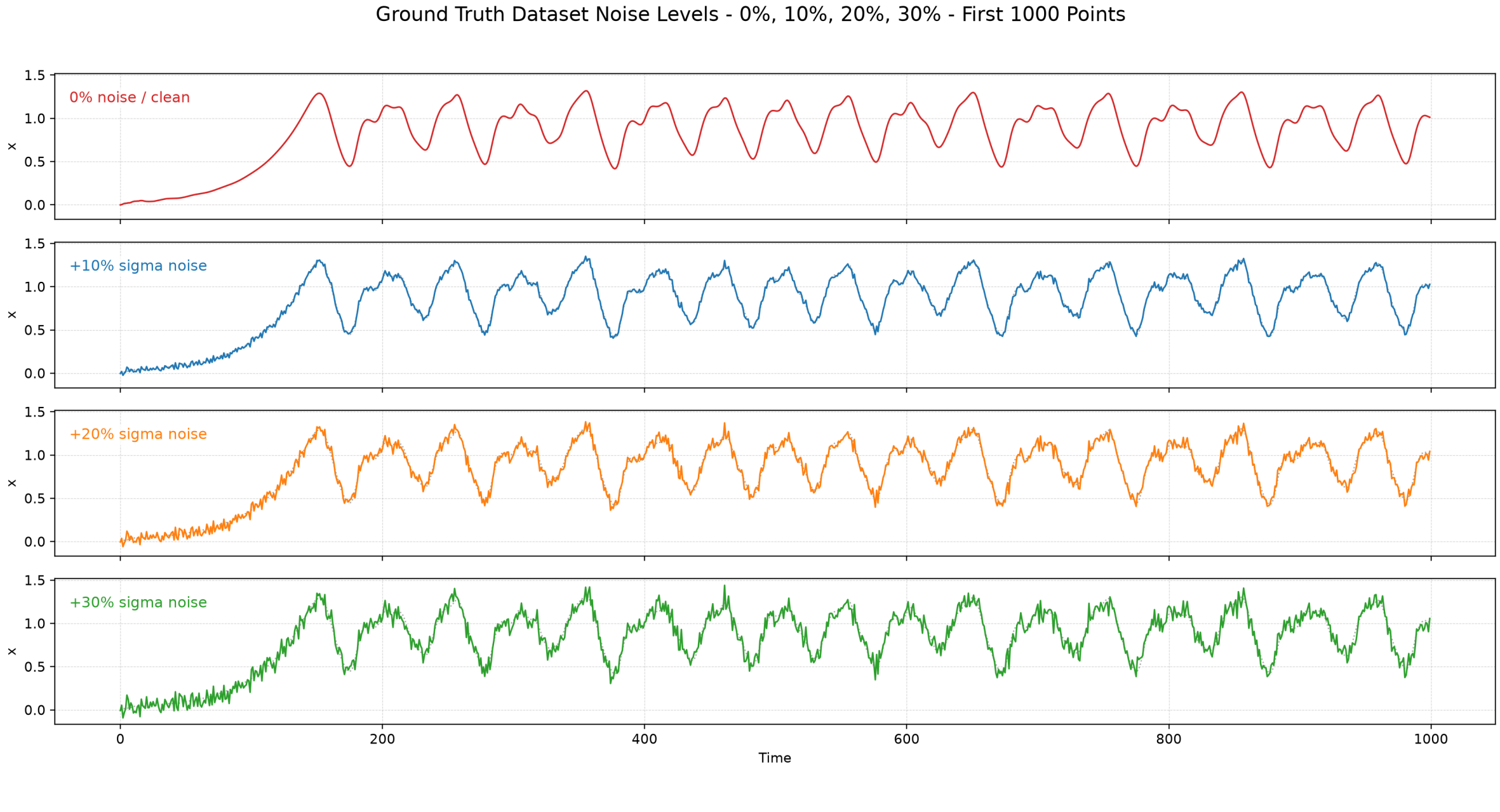

(a)

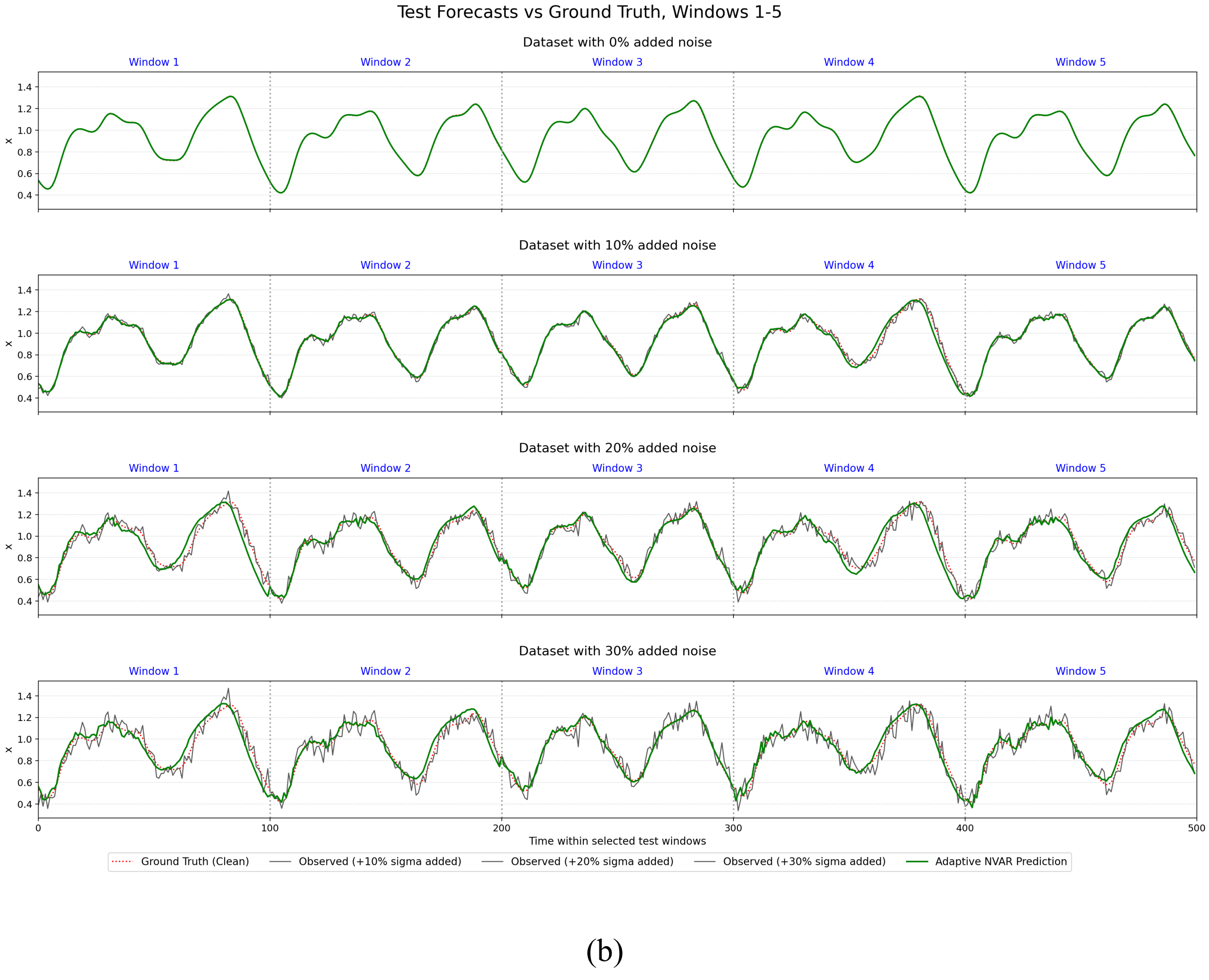

(b)

**Figure 5**. (a) Mackey-Glass dataset with chaotic parameter settings and different levels of additive Gaussian noise (0%, 10%, 20%, 30%). (b) Forecast by Adaptive NVAR with CheMLFlow on 5 independent windows of the Mackey-Glass dataset with 0%, 10%, 20%, 30% additive Gaussian noise.

To assess the robustness of the model, different levels of synthetic Gaussian noise are added to the already chaotic dataset. **Table S1** reports the forecasting results for noise levels of 0-30% and for different forecasting horizons, namely 25, 50, 75, and 100 prediction steps. The results are then compared with those originally reported in the Adaptive NVAR paper, showing that CheMLFlow successfully forecasts this chaotic time series with performance comparable to the reported benchmark. Future work will focus on applying CheMLFlow's time-series forecasting module to various chemistry datasets. We also plan to extend the current implementation by testing additional time series models and comparing their performance across both synthetic chaotic systems and real chemical datasets.

# 4. CONCLUSIONS

We have developed CheMLFlow, an open-source, configuration driven software for building, benchmarking, and applying reproducible and explainable molecular ML/AI workflows in a high throughput fashion. This work demonstrated CheMLFlow as a chemistry aware, agent-assisted ML/AI workflow development and experimentation platform. CheMLFlow was applied for bioactivity predictions with selected model combinations and for benchmarking models of molecular datasets for predicting quantum mechanical, physicochemical, fuel performance, and ADME properties, where leaderboard like reports were obtained through a single DOE-driven approach. Further, we showed the AI agent-assisted capabilities for odor prediction tasks on QuantumScents dataset, and beyond molecular datasets, a time series forecasting module for chaotic dynamical systems.

A central contribution of CheMLFlow is not a single model or descriptor, but a way to organize the surrounding scientific process. Data ingestion, curation, representation, splitting, preprocessing, model training, evaluation, and analysis are treated as explicit pipeline nodes with defined artifacts. The benchmark studies showed why these matter as performance can shift with the representation, split strategy, preprocessing choices, and model family, and failures can be informative rather than merely discarded. By expanding these choices into DOE runs, CheMLFlow makes each result traceable to a concrete configuration while allowing successful runs to be aggregated and failed runs to remain visible for audit. A second implication is that the resulting artifact set is useful beyond static benchmarking. Configurations, split files, metrics, predictions, failure records, and analysis summaries form structured context that can be inspected by people reproducing the experiment and by agent-assisted workflows. An agent could use this evidence to suggest follow-up experiments, flag unstable workflow regions, or draft reports from recorded results. Future work will extend the supported datasets, model families, reporting outputs, and automation patterns. While current work applied CheMLFlow to study materials properties for small molecule data sets, future works aim to extending them for periodic systems truly reaching the potential of materials informatics. Finally, CheMLFlow is being developed with practical applications in mind and is deliberately workflow centered focusing on reducing orchestration burden, preserving provenance, and making comparisons across complete scientific ML pipelines easier, while ensuring the pipelines are easy to reproduce and evaluate.

# 5. SOFTWARE AND DATA AVAILABILITY

CheMLFlow is available at https://github.com/nijamudheen/CheMLFlow.

# 6. AUTHOR CONTRIBUTIONS

N.A. developed the original idea. B.S., S.L-M., E.D-C., and N.A. conceptualized the ideas, implemented the software, designed and run the experiments, and analyzed the results. B.S. and N.A. wrote the manuscript with contributions from S.L.M. and E.D.C. S.K. supervised the implementation of Adaptive NVAR model. All authors have read and approved the final version of the manuscript. N.A. and J.L.M.C. coordinated the project meetings and discussions. J.L.M.C. acquired funding for running part of the experiments in HPCC at ICER, MSU.

# 7. CONFLICTS OF INTERESTS

B.S. and N.A. are part of an early stage startup Kernfield Labs Ltd. based in London, UK. Other authors declare no conflicts of interests.

# 8. ACKNOWLEDGEMENTS

This research received no specific grant from any funding agency in the public, commercial, or not-for-profit sectors. We are grateful for the computational support from the Institute for Cyber-Enabled Research (ICER) at the Michigan State University. S.L.M., E.D.C., and S.K. are grateful for the support from the National Research Foundation (NRF) of Korea grant funded by the Korea government (MSIT) (No. 2022R1A5A1033624, RS-2024-00406152, RS-2025-00517727) and Global Learning & Academic research institution for Masters, Ph.D. students, and Postdocs (LAMP) by the Ministry of Education (No. RS-2023-00301938).

# 9. ORCID

Brendan Smith: https://orcid.org/0000-0003-3460-9984

Susana López-Moreno: https://orcid.org/0000-0002-1322-2775

Eric Dolores-Cuenca: https://orcid.org/0000-0003-3909-5580

Sangil Kim: https://orcid.org/0000-0002-4408-2904

Jose L. Mendoza-Cortes: https://orcid.org/0000-0001-5184-1406

Nijamudheen Abdulrahiman: https://orcid.org/0000-0001-9191-1851

# 10. CORRESPONDING AUTHOR

Jose L. Mendoza-Cortes: jmendoza@msu.edu

Nijamudheen Abdulrahiman: apchnijam@gmail.com, nijam@kernfieldlabs.com

# 11. REFERENCES

(1) Seal, S.; Mahale, M.; García-Ortegón, M.; Joshi, C. K.; Hosseini-Gerami, L.; Beatson, A.; Greenig, M.; Shekhar, M.; Patra, A.; Weis, C.; Mehrjou, A.; Badré, A.; Paisley, B.; Lowe, R.; Singh, S.; Shah, F.; Johannesson, B.; Williams, D.; Rouquie, D.; Clevert, D.-A.; Schwab, P.; Richmond, N.; Nicolaou, C. A.; Gonzalez, R. J.; Naven, R.; Schramm, C.; Vidler, L. R.; Mansouri, K.; Walters, W. P.; Wilk, D. D.; Spjuth, O.; Carpenter, A. E.; Bender, A. Machine Learning for Toxicity Prediction Using Chemical Structures: Pillars for Success in the Real World. *Chem. Res. Toxicol.* **2025**. https://doi.org/10.1021/acs.chemrestox.5c00033.

(2) Tom, G.; Schmid, S. P.; Baird, S. G.; Cao, Y.; Darvish, K.; Hao, H.; Lo, S.; Pablo-García, S.; Rajaonson, E. M.; Skreta, M.; Yoshikawa, N.; Corapi, S.; Akkoc, G. D.; Strieth-Kalthoff, F.; Seifrid, M.; Aspuru-Guzik, A. Self-Driving Laboratories for Chemistry and Materials Science. *Chemical Reviews*. American Chemical Society August 28, 2024, pp 9633–9732. https://doi.org/10.1021/acs.chemrev.4c00055.

(3) Choudhary, K.; DeCost, B.; Chen, C.; Jain, A.; Tavazza, F.; Cohn, R.; Park, C. W.; Choudhary, A.; Agrawal, A.; Billinge, S. J. L.; Holm, E.; Ong, S. P.; Wolverton, C. Recent Advances and Applications of Deep Learning Methods in Materials Science. *NPJ Comput. Mater.* **2022**, *8* (1), 59. https://doi.org/10.1038/s41524-022-00734-6.

(4) Sadybekov, A. V.; Katritch, V. Computational Approaches Streamlining Drug Discovery. *Nature* **2023**, *616* (7958), 673–685. https://doi.org/10.1038/s41586-023-05905-z.

(5) Gorgulla, C. Recent Developments in Ultralarge and Structure-Based Virtual Screening Approaches. *Annual Review of Biomedical Data Science Downloaded from www.annualreviews.org. Guest* **2026**. https://doi.org/10.1146/annurev-biodatasci-020222.

(6) Scannell, J. W.; Bosley, J.; Hickman, J. A.; Dawson, G. R.; Truebel, H.; Ferreira, G. S.; Richards, D.; Treherne, J. M. Predictive Validity in Drug Discovery: What It Is, Why It Matters and How to Improve It. *Nat. Rev. Drug Discov.* **2022**, *21* (12), 915–931. https://doi.org/10.1038/s41573-022-00552-x.

(7) Tu, Z.; Stuyver, T.; Coley, C. W. Predictive Chemistry: Machine Learning for Reaction Deployment, Reaction Development, and Reaction Discovery. *Chemical Science*. Royal Society of Chemistry November 28, 2022, pp 226–244. https://doi.org/10.1039/d2sc05089g.

(8) Lombardo, T.; Duquesnoy, M.; El-Bouysidy, H.; Årén, F.; Gallo-Bueno, A.; Jørgensen, P. B.; Bhowmik, A.; Demortière, A.; Ayerbe, E.; Alcaide, F.; Reynaud, M.; Carrasco, J.; Grimaud, A.; Zhang, C.; Vegge, T.; Johansson, P.; Franco, A. A. Artificial Intelligence Applied to Battery Research: Hype or Reality? *Chem. Rev.* **2022**, *122* (12), 10899–10969. https://doi.org/10.1021/acs.chemrev.1c00108.

(9) Yao, Z.; Lum, Y.; Johnston, A.; Mejia-Mendoza, L. M.; Zhou, X.; Wen, Y.; Aspuru-Guzik, A.; Sargent, E. H.; Seh, Z. W. Machine Learning for a Sustainable Energy Future. *Nat. Rev. Mater.* **2022**, *8* (3), 202–215. https://doi.org/10.1038/s41578-022-00490-5.

(10) Bozal-Ginesta, C.; Pablo-García, S.; Choi, C.; Tarancón, A.; Aspuru-Guzik, A. Developing Machine Learning for Heterogeneous Catalysis with Experimental and Computational Data. *Nat. Rev. Chem.* **2025**, *9* (9), 601–616. https://doi.org/10.1038/s41570-025-00740-4.

(11) Tropsha, A.; Isayev, O.; Varnek, A.; Schneider, G.; Cherkasov, A. Integrating QSAR Modelling and Deep Learning in Drug Discovery: The Emergence of Deep QSAR. *Nat. Rev. Drug Discov.* **2024**, *23* (2), 141–155. https://doi.org/10.1038/s41573-023-00832-0.

(12) Zhang, O.; Lin, H.; Zhang, X.; Wang, X.; Wu, Z.; Ye, Q.; Zhao, W.; Wang, J.; Ying, K.; Kang, Y.; Hsieh, C.-Y.; Hou, T. Graph Neural Networks in Modern AI-Aided Drug Discovery. *Chem. Rev.* **2025**, *125* (20), 10001–10103. https://doi.org/10.1021/acs.chemrev.5c00461.

(13) Pyzer-Knapp, E. O.; Manica, M.; Staar, P.; Morin, L.; Ruch, P.; Laino, T.; Smith, J. R.; Curioni, A. Foundation Models for Materials Discovery – Current State and Future Directions. *NPJ Comput. Mater.* **2025**, *11* (1), 61. https://doi.org/10.1038/s41524-025-01538-0.

(14) Choi, J.; Nam, G.; Choi, J.; Jung, Y. A Perspective on Foundation Models in Chemistry. *JACS Au* **2025**, *5* (4), 1499–1518. https://doi.org/10.1021/jacsau.4c01160.

(15) Song, B.; Zhang, J.; Liu, Y.; Liu, Y.; Jiang, J.; Yuan, S.; Zhen, X.; Liu, Y. A Systematic Review of Molecular Representation Learning Foundation Models. *Briefings in Bioinformatics*. Oxford University Press January 1, 2026. https://doi.org/10.1093/bib/bbaf703.

(16) Tang, X.; Dai, H.; Knight, E.; Wu, F.; Li, Y.; Li, T.; Gerstein, M. A Survey of Generative AI for de Novo Drug Design: New Frontiers in Molecule and Protein Generation. *Briefings in Bioinformatics*. Oxford University Press July 1, 2024. https://doi.org/10.1093/bib/bbae338.

(17) M. Bran, A.; Cox, S.; Schilter, O.; Baldassari, C.; White, A. D.; Schwaller, P. Augmenting Large Language Models with Chemistry Tools. *Nat. Mach. Intell.* **2024**, *6* (5), 525–535. https://doi.org/10.1038/s42256-024-00832-8.

(18) Boiko, D. A.; MacKnight, R.; Kline, B.; Gomes, G. Autonomous Chemical Research with Large Language Models. *Nature* **2023**, *624* (7992), 570–578. https://doi.org/10.1038/s41586-023-06792-0.

(19) Gottweis, J.; Weng, W.-H.; Daryin, A.; Tu, T.; Sirkovic, P.; Myaskovsky, A.; Glowaty, G.; Weissenberger, F.; Orlandi, A.; Popovici, D.; Palepu, A.; Rong, K.; Tanno, R.; Saab, K.; Zhang, F.; Blum, J.; Carroll, A.; Kulkarni, K.; Tomašev, N.; Zverinski, D.; Rendulic, I.; Vedadi, E.; Hasler, F.; Rimanic, L.; Boia, M.; Budiselic, I.; Feinstein, B.; Bellaiche, M.; Sheffer, T.; Freyberg, J.; Ratcliff, J.; Bertolli, O.; Chou, K.; Hassidim, A.; Gokturk, B.; Vahdat, A.; Guan, Y.; Dhillon, V.; Vaishnav, E. D.; Lee, B.; Costa, T. R. D.; Penadés,

J. R.; Peltz, G.; Matias, Y.; Manyika, J.; Hassabis, D.; Xu, Y.; Kohli, P.; Pawlosky, A.; Karthikesalingam, A.; Natarajan, V. Accelerating Scientific Discovery with Co-Scientist. *Nature* **2026**, *655* (8122), 487–496. https://doi.org/10.1038/s41586-026-10644-y.

(20) Aspuru-Guzik, A.; Bernales, V. The Rise of Agents: Computational Chemistry Is Ready for (R)Evolution. *Polyhedron* **2025**, *281*. https://doi.org/10.1016/j.poly.2025.117707.

(21) Zou, Y.; Cheng, A. H.; Aldossary, A.; Bai, J.; Leong, S. X.; Campos-Gonzalez-Angulo, J. A.; Choi, C.; Ser, C. T.; Tom, G.; Wang, A.; Zhang, Z.; Yakavets, I.; Hao, H.; Crebolder, C.; Bernales, V.; Aspuru-Guzik, A. El Agente: An Autonomous Agent for Quantum Chemistry. *Matter* **2025**, *8* (7), 102263. https://doi.org/10.1016/j.matt.2025.102263.

(22) Ramos, M. C.; Collison, C. J.; White, A. D. A Review of Large Language Models and Autonomous Agents in Chemistry. *Chem. Sci.* **2025**, *16* (6), 2514–2572. https://doi.org/10.1039/D4SC03921A.

(23) Özçelik, R.; Brinkmann, H.; Criscuolo, E.; Grisoni, F. Generative Deep Learning for de Novo Drug Design─A Chemical Space Odyssey. *Journal of Chemical Information and Modeling*. American Chemical Society July 28, 2025, pp 7352–7372. https://doi.org/10.1021/acs.jcim.5c00641.

(24) Yang, K.; Swanson, K.; Jin, W.; Coley, C.; Eiden, P.; Gao, H.; Guzman-Perez, A.; Hopper, T.; Kelley, B.; Mathea, M.; Palmer, A.; Settels, V.; Jaakkola, T.; Jensen, K.; Barzilay, R. Analyzing Learned Molecular Representations for Property Prediction. *J. Chem. Inf. Model.* **2019**, *59* (8), 3370–3388. https://doi.org/10.1021/acs.jcim.9b00237.

(25) Heid, E.; Greenman, K. P.; Chung, Y.; Li, S. C.; Graff, D. E.; Vermeire, F. H.; Wu, H.; Green, W. H.; McGill, C. J. Chemprop: A Machine Learning Package for Chemical Property Prediction. *J. Chem. Inf. Model.* **2024**, *64* (1), 9–17. https://doi.org/10.1021/acs.jcim.3c01250.

(26) Loeffler, H. H.; He, J.; Tibo, A.; Janet, J. P.; Voronov, A.; Mervin, L. H.; Engkvist, O. Reinvent 4: Modern AI–Driven Generative Molecule Design. *J. Cheminform.* **2024**, *16* (1). https://doi.org/10.1186/s13321-024-00812-5.

(27) Zdrazil, B.; Felix, E.; Hunter, F.; Manners, E. J.; Blackshaw, J.; Corbett, S.; de Veij, M.; Ioannidis, H.; Lopez, D. M.; Mosquera, J. F.; Magarinos, M. P.; Bosc, N.; Arcila, R.; Kizilören, T.; Gaulton, A.; Bento, A. P.; Adasme, M. F.; Monecke, P.; Landrum, G. A.; Leach, A. R. The ChEMBL Database in 2023: A Drug Discovery Platform Spanning Multiple Bioactivity Data Types and Time Periods. *Nucleic Acids Res.* **2024**, *52* (D1), D1180–D1192. https://doi.org/10.1093/nar/gkad1004.

(28) Gawalska, A.; Czub, N.; Sapa, M.; Kołaczkowski, M.; Bucki, A.; Mendyk, A. Application of Automated Machine Learning in the Identification of Multi-target-directed Ligands Blocking PDE4B, PDE8A, and TRPA1 with Potential Use in the Treatment of Asthma and COPD. *Mol. Inform.* **2023**, *42* (7). https://doi.org/10.1002/minf.202200214.

(29) Nada, H.; Gul, A. R.; Elkamhawy, A.; Kim, S.; Kim, M.; Choi, Y.; Park, T. J.; Lee, K. Machine Learning-Based Approach to Developing Potent EGFR Inhibitors for Breast Cancer─Design, Synthesis, and In Vitro Evaluation. *ACS Omega* **2023**, *8* (35), 31784–31800. https://doi.org/10.1021/acsomega.3c02799.

(30) Sanches, I. H.; Braga, R. C.; Alves, V. M.; Andrade, C. H. Enhancing HERG Risk Assessment with Interpretable Classificatory and Regression Models. *Chem. Res. Toxicol.* **2024**, *37* (6), 910–922. https://doi.org/10.1021/acs.chemrestox.3c00400.

(31) Espinoza, G. Z.; Angelo, R. M.; Oliveira, P. R.; Honorio, K. M. Evaluating Deep Learning Models for Predicting ALK-5 Inhibition. *PLoS One* **2021**, *16* (1), e0246126. https://doi.org/10.1371/journal.pone.0246126.

(32) Lipinski, C. A.; Lombardo, F.; Dominy, B. W.; Feeney, P. J. Experimental and Computational Approaches to Estimate Solubility and Permeability in Drug Discovery and Development Settings. *Adv. Drug Deliv. Rev.* **1997**, *23* (1–3), 3–25. https://doi.org/10.1016/S0169-409X(96)00423-1.

(33) Bickerton, G. R.; Paolini, G. V.; Besnard, J.; Muresan, S.; Hopkins, A. L. Quantifying the Chemical Beauty of Drugs. *Nat. Chem.* **2012**, *4* (2), 90–98. https://doi.org/10.1038/nchem.1243.

(34) Landrum, G. RDKit: Open-Source Cheminformatics. Https://Www.Rdkit.Org. 2020.

(35) Ramakrishnan, R.; Dral, P. O.; Rupp, M.; von Lilienfeld, O. A. Quantum Chemistry Structures and Properties of 134 Kilo Molecules. *Sci. Data* **2014**, *1* (1), 140022. https://doi.org/10.1038/sdata.2014.22.

(36) Zhou, G.; Gao, Z.; Ding, Q.; Zheng, H.; Xu, H.; Wei, Z.; Zhang, L.; Ke, G. Uni-Mol: A Universal 3D Molecular Representation Learning Framework. March 6, 2023. https://doi.org/10.26434/chemrxiv-2022-jjm0j-v4.

(37) Broccatelli, F.; Carosati, E.; Neri, A.; Frosini, M.; Goracci, L.; Oprea, T. I.; Cruciani, G. A Novel Approach for Predicting P-Glycoprotein (ABCB1) Inhibition Using Molecular Interaction Fields. *J. Med. Chem.* **2011**, *54* (6), 1740–1751. https://doi.org/10.1021/jm101421d.

(38) Stratiichuk, R.; Shevchuk, N.; Kyrylenko, R.; Vozniak, V.; Koleiev, I.; Voitsitskyi, T.; Korogodski, S.; Ostrovsky, Z.; Khropachov, I.; Vasylevskyi, V.; Husak, V.; Starosyla, S.; Yesylevskyy, S.; Nafiiev, A. Improving ADMET Prediction with Descriptor Augmentation of Mol2Vec Embeddings. July 18, 2025. https://doi.org/10.1101/2025.07.14.664363.

(39) Schaduangrat, N.; Anuwongcharoen, N.; Charoenkwan, P.; Shoombuatong, W. DeepAR: A Novel Deep Learning-Based Hybrid Framework for the Interpretable Prediction of Androgen Receptor Antagonists. *J. Cheminform.* **2023**, *15* (1). https://doi.org/10.1186/s13321-023-00721-z.

(40) Saldana, D. A.; Starck, L.; Mougin, P.; Rousseau, B.; Pidol, L.; Jeuland, N.; Creton, B. Flash Point and Cetane Number Predictions for Fuel Compounds Using Quantitative

Structure Property Relationship (QSPR) Methods. *Energy & Fuels* **2011**, *25* (9), 3900–3908. https://doi.org/10.1021/ef200795j.

(41) Das, D. D.; St. John, P. C.; McEnally, C. S.; Kim, S.; Pfefferle, L. D. Measuring and Predicting Sooting Tendencies of Oxygenates, Alkanes, Alkenes, Cycloalkanes, and Aromatics on a Unified Scale. *Combust. Flame* **2018**, *190*, 349–364. https://doi.org/10.1016/j.combustflame.2017.12.005.

(42) Burns, J. W.; Green, W. H. Generalizable, Fast, and Accurate DeepQSPR with Fastprop. *J. Cheminform.* **2025**, *17* (1). https://doi.org/10.1186/s13321-025-01013-4.

(43) Arockiaraj, M.; Paul, D.; Clement, J.; Tigga, S.; Jacob, K.; Balasubramanian, K. Novel Molecular Hybrid Geometric-Harmonic-Zagreb Degree Based Descriptors and Their Efficacy in QSPR Studies of Polycyclic Aromatic Hydrocarbons. *SAR QSAR Environ. Res.* **2023**, *34* (7), 569–589. https://doi.org/10.1080/1062936X.2023.2239149.

(44) Burns, J. W.; Zalte, A. S.; Abreu, C. R. A.; Sieg, J.; Feldmann, C.; Mathea, M.; Green, W. H. Deep Learning Foundation Models from Classical Molecular Descriptors. **2026**.

(45) Akiba, T.; Sano, S.; Yanase, T.; Ohta, T.; Koyama, M. Optuna. In *Proceedings of the 25th ACM SIGKDD International Conference on Knowledge Discovery & Data Mining*; ACM: New York, NY, USA, 2019; pp 2623–2631. https://doi.org/10.1145/3292500.3330701.

(46) Burns, J. W.; Rogers, D. M. QuantumScents: Quantum-Mechanical Properties for 3.5k Olfactory Molecules. *J. Chem. Inf. Model.* **2023**, *63* (23), 7330–7337. https://doi.org/10.1021/acs.jcim.3c01338.

(47) Sherkhon, A.; Lopez-Moreno, S.; Dolores-Cuenca, E.; Lee, S.; Kim, S. Adaptive Nonlinear Vector Autoregression: Robust Forecasting for Noisy Chaotic Time Series. **2025**.

# Supporting Information for

# CheMLFlow: An Open-Source Platform for Cheminformatics and Materials Informatics Applications

Brendan Smith[1], Susana López-Moreno[2,3,4], Eric Dolores-Cuenca[5], Sangil Kim[2,4], Jose L. Mendoza-Cortes[6,7],* Nijamudheen Abdulrahiman[1*]

[1]Kernfield Labs, London, United Kingdom.
[2]Department of Mathematics, Pusan National University, Republic of Korea.
[3] Humanoid Olfactory Display Center, Pusan National University, Republic of Korea.
[4] Industrial Mathematics Center, Pusan National University, Republic of Korea.
[5] Yonsei University, Republic of Korea.
[6] Department of Chemical Engineering & Materials Science, Michigan State University, United States.
[7] Department of Physics & Astronomy, Michigan State University, East Lansing, Michigan 48824, United States.

# 1. Adaptive NVAR Model Implementation

Adaptive NVAR is a machine-learning algorithm for time-series forecasting with a simple architecture that improves scalability and robustness to noisy data. The model architecture is shown in Figure S1. The benchmarking results are summarized in Table S1.

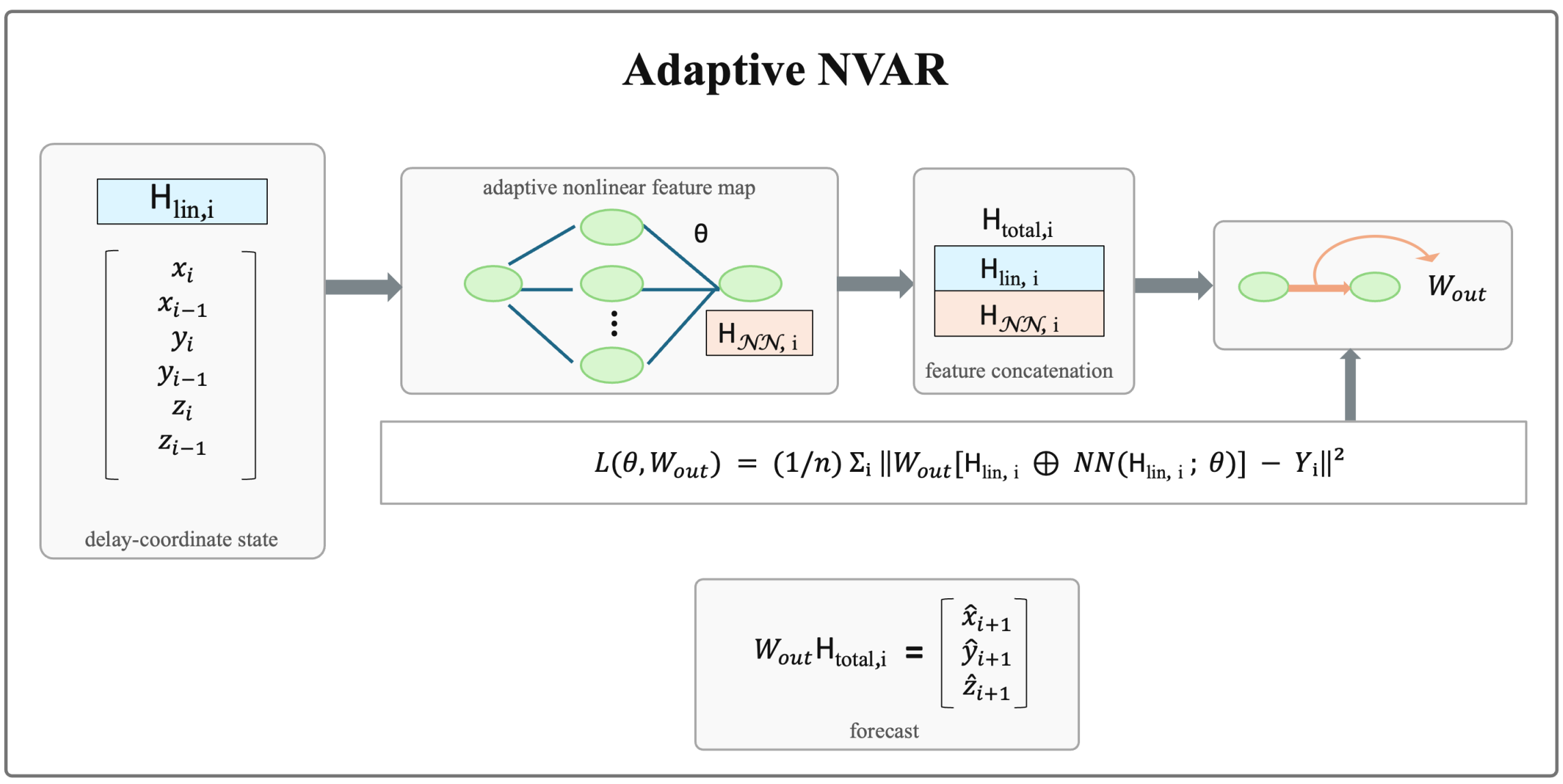

**Figure S1**: Adaptive NVAR model implemented in CheMLFlow.

**Table S1**. DOE-guided benchmarking of Adaptive NVAR with CheMLFlow on the Mackey-Glass dataset for different amounts of noise (0%, 10%, 20%, 30%) and different forecasting horizons (25, 50, 75, 100 steps).

| **Noise** | **Forecasting horizons** | **Perf** (RMSE) | **SOTA** |
|---|---|---|---|
| | | | **Perf** |

| 0% | 25 | 0.002 | 0.002 |
|---|---|---|---|
| | 50 | 0.003 | 0.004 |
| | 75 | 0.005 | 0.005 |
| | 100 | 0.007 | 0.006 |
| 10% | 25 | 0.009 | 0.010 |
| | 50 | 0.011 | 0.011 |
| | 75 | 0.014 | 0.013 |
| | 100 | 0.017 | 0.016 |
| 20% | 25 | 0.018 | 0.020 |
| | 50 | 0.022 | 0.024 |
| | 75 | 0.026 | 0.028 |
| | 100 | 0.032 | 0.032 |
| 30% | 25 | 0.028 | 0.029 |
| | 50 | 0.032 | 0.036 |
| | 75 | 0.039 | 0.043 |
| | 100 | 0.046 | 0.051 |